%% file: main.tex
\documentclass[11pt,a4paper]{article}
\usepackage[hyperref]{naaclhlt2019}
\usepackage{times}
\usepackage{latexsym}

\usepackage{url}

\aclfinalcopy 
\def\aclpaperid{642}  

\usepackage{helvet}  
\usepackage{courier}  
\usepackage{graphicx}  
\usepackage{booktabs}
\usepackage{enumitem}
\usepackage{amsmath}
\usepackage{amsfonts}
\usepackage{comment}
\usepackage{shadowtext}
\usepackage{xspace}
\newcommand{\ReCoRD}{{\fontfamily{fla}\fontseries{m}\selectfont Re}{\fontfamily{fla}\fontseries{b}\selectfont{Co}}{\fontfamily{fla}\fontseries{m}\selectfont RD}\xspace}

\usepackage{todonotes} 
\tikzset{inlinenotestyle/.append style={align=justify}}

\title{Unsupervised Deep Structured Semantic Models \\ for Commonsense Reasoning}
\author{ Shuohang Wang$^1$\thanks{~~Work done when the author was at Microsoft}, Sheng Zhang$^2$, Yelong Shen$^4$, Xiaodong Liu$^3$, \\ 
{ \bf  Jingjing Liu$^3$, Jianfeng Gao$^3$, Jing Jiang$^1$} \\
$^1${Singapore Management University},$^2${Johns Hopkins University}, $^3${Microsoft}, $^4${Tencent AI Lab}\\
{\texttt {\{shwang.2014,jingjiang\}@smu.edu.sg, zsheng2@jhu.edu}} \\
{\texttt {\{xiaodl,jingjl,jfgao\}@microsoft.com, yelongshen@tencent.com } }
}
 
\begin{document}
%
\maketitle
\begin{abstract}

Commonsense reasoning is fundamental to natural language understanding. While traditional methods rely heavily on human-crafted features and knowledge bases, we explore learning commonsense knowledge from a large amount of raw text via unsupervised learning. We propose two neural network models based on the Deep Structured Semantic Models (DSSM) framework to tackle two classic commonsense reasoning tasks, Winograd Schema challenges (WSC) and Pronoun Disambiguation (PDP).  Evaluation shows that the proposed models effectively capture contextual information in the sentence and co-reference information between pronouns and nouns, and achieve significant improvement over previous state-of-the-art approaches. 
\end{abstract}
\input{intro}
\input{related}

\input{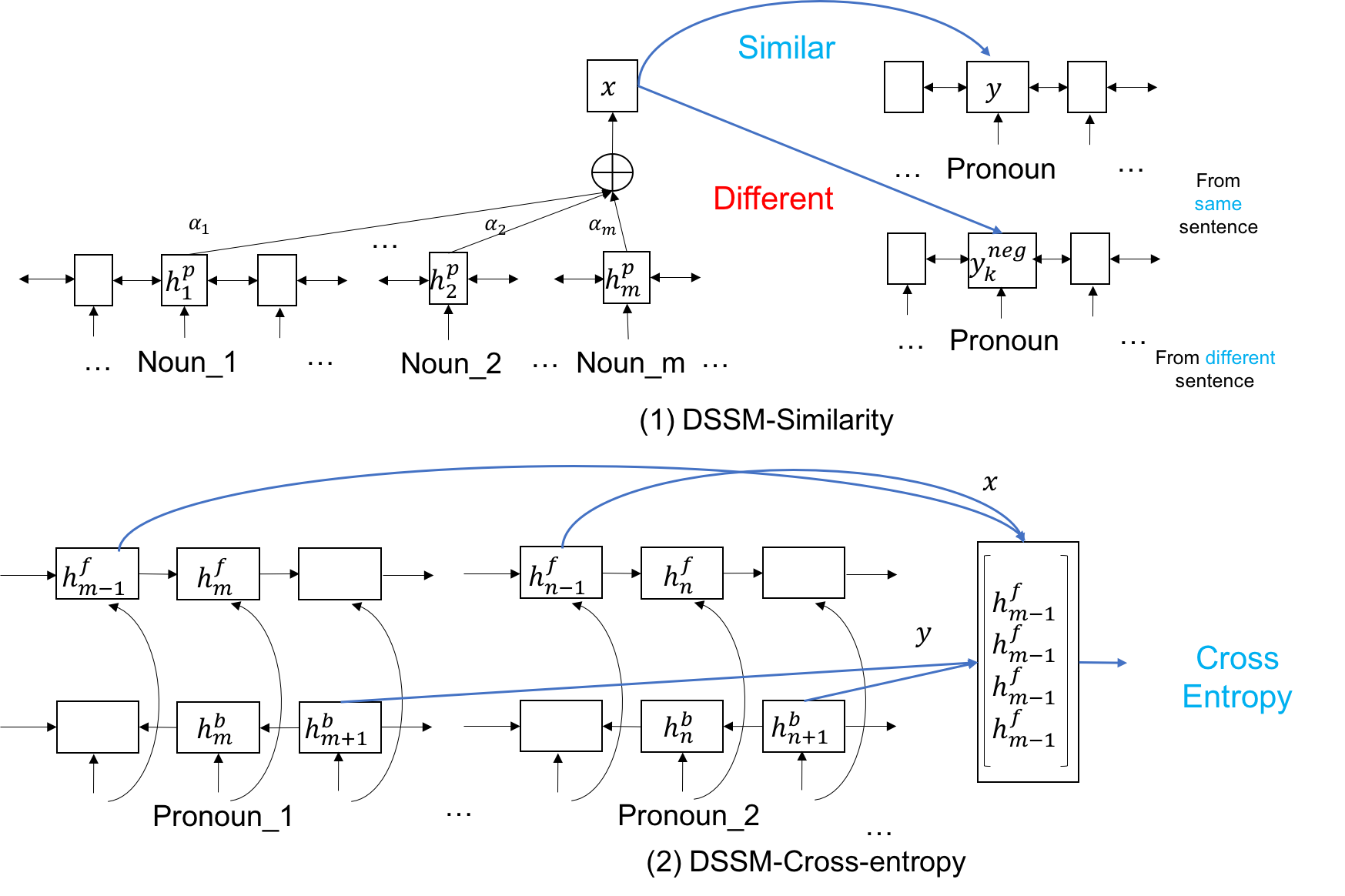}

\input{exp}

\section{Conclusion}
In conclusion, to overcome the lack of human labeled data, we proposed two unsupervised deep structured semantic models (UDSSM) for commonsense reasoning.
We evaluated our models on the commonsense reasoning tasks of Pronoun Disambiguation Problems (PDP) and Winograd Schema Challenge~\cite{levesque2011winograd}, where the questions are quite easy for human to answer, but quite challenging for the machine.
Without using any hand-craft knowledge base, our model achieved stat-of-the-art performance on the two tasks.

In the future work, we will use Transformer, which is proved to be more powerful than LSTM, as the encoder of our unsupervised deep structured semantic models, and we will collect a larger corpus from Common Crawl to train our model.
\bibliography{naaclhlt2019}
\bibliographystyle{acl_natbib}
\end{document}

%% file: intro.tex
\section{Introduction}
\label{sect:intro}

Commonsense reasoning 
is concerned with simulating the human ability to make presumptions about the type and essence of ordinary situations they encounter every day~\cite{davis2015commonsense}. 
It is one of the key challenges in natural language understanding, and has drawn increasing attention in recent years~\cite{levesque2011winograd,roemmele2011choice,zhang2016ordinal,rashkin2018modeling,rashkin2018event2mind,zellers2018swag,trinh2018simple}. 
However, due to the lack of labeled training data or 
comprehensive hand-crafted knowledge bases, commonsense reasoning tasks such as Winograd Schema Challenge~\cite{levesque2011winograd} are still far from being solved.

\begin{table}[!ht]

\begin{enumerate} [leftmargin=*]

\item \textit{The city councilmen refused the demonstrators a permit because \textbf{they} feared violence.} 

Who feared violence? \\
A. \textbf{The city councilmen}\hspace{0.02\textwidth} B. The demonstrators

\item \textit{The city councilmen refused the demonstrators a permit because \textbf{they} advocated violence.} 
\vspace{0.1cm}\\
Who advocated violence? \\
A. The city councilmen\hspace{0.02\textwidth} B. \textbf{The demonstrators}

\end{enumerate}
\caption{Examples from Winograd Schema Challenge (WSC). The task is to identify the reference of the pronoun in bold.}
\label{tbl:example}
\end{table}





In this work, we propose two effective unsupervised models for commonsense reasoning, and evaluate them on two classic commonsense reasoning tasks: Winograd Schema Challenge (WSC) and Pronoun Disambiguation Problems (PDP).
Compared to other commonsense reasoning tasks, WSC and PDP better approximate real human reasoning, and can be more easily solved by native English-speaking adults~\cite{levesque2011winograd}. 
In addition, they are also technically challenging. 
For example, the best reported result on WSC is only 20 percentage points better than random guess in accuracy~\cite{radford2019language}.

Table~\ref{tbl:example} shows two examples from WSC.
In order to resolve the co-reference in these two examples, one cannot predict what ``\textbf{\textit{they}}'' refers to unless she is equipped with the commonsense knowledge that ``\textit{demonstrators usually cause violence and city councilmen usually fear violence}''.



As no labeled training data is available for these tasks, previous approaches are based on either hand-crafted knowledge bases or large-scale language models. 
For example, \citet{liu2016combing} used existing knowledge bases such as ConceptNet~\cite{liu2004conceptnet} and WordNet~\cite{miller1995wordnet} for external supervision to train word embeddings and solve the WSC challenge.
Recently, \citet{trinh2018simple} first used raw text from books/news to train a neural Language Model (LM), and then employed the trained model to compare the probabilities of the sequences, where the pronouns are replaced by each of the candidate references, and to pick the candidate that leads to the highest probability as the answer.

Because none of the existing hand-crafted knowledge bases is comprehensive enough 
to cover all the world knowledge\footnote{We don't believe it is possible to construct such a knowledge base given that the world is changing constantly.}, 
we focus on building unsupervised models that can learn commonsense knowledge directly from unlimited raw text. Different from the neural language models, our models are optimized for co-reference resolution and achieve much better results on both the PDP and WSC tasks.

In this work we formulate the two commonsense reasoning tasks in WSC and PDP as a pairwise ranking problem. 
As the first example in Table~\ref{tbl:example}, we want to develop a pair-wise scoring model $\text{Score}_{\theta}(x_i,y)$ that scores the correct antecedent-pronoun pair (``\textit{councilmen}``, ``\textit{they}'') higher than the incorrect one (``\textit{demonstrators}``, ``\textit{they}''). 
These scores depend to a large degree upon the contextual information of the pronoun (e.g., “they”) and the candidate antecedent (e.g., “councilmen”). 
In other words, it requires to capture the semantic meaning of the pronoun and the candidate antecedent based on the sentences where they occur, respectively. 

\begin{figure*}[!ht]
\centering
\includegraphics[width=1\textwidth]{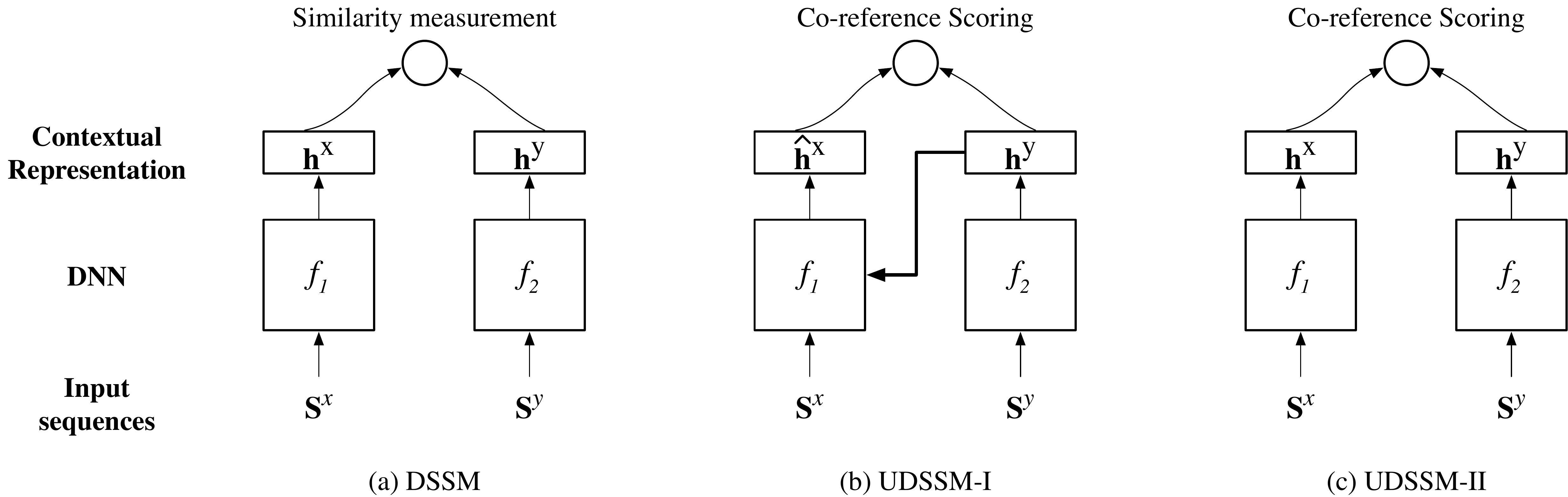}
\caption{An overview of (a) the general framework of Deep Structured Semantic Model (DSSM) and our two unsupervised models based on DSSM: (b) UDSSM-I and (c) UDSSM-II. 
Compared with DSSM, both UDSSM-I and UDSSM-II compute co-reference scores instead of similarity.
\label{fig:dssm}}
\end{figure*}

To tackle this issue, we propose two models based on the framework of Deep Structured Similarity Model~(DSSM)~\cite{huang2013learning}, as shown in Figure~\ref{fig:dssm}(a). 
Formally, let $S^x$ be the sentence containing the candidate antecedent $x_i$ and $S^y$ the sentence containing the pronoun y which we're interested in.”
DSSM measures the semantic similarity of a pair of inputs $(x_i,y)$ by 1) mapping $x_i$ and $y$, together with their context information, into two vectors in a semantic space using deep neural networks $f_1$ and $f_2$, parameterized by $\theta$; and 2) computing cosine similarity\footnote{DSSMs can be applied to a wide range of tasks depending on the definition of $(x, y)$. For example, $(x, y)$ is a query-document pair for Web search ranking, a document pair in recommendation, a question-answer pair in QA, and so on. See Chapter 2 of \cite{gaosurvey} for a survey.} between them. In our case, we need to learn a task-specific semantic space where the distance between two vectors measures how likely they co-refer. 
Commonsense knowledge such as ``demonstrators usually cause violence'' can be implicitly captured in the semantic space through DSSM, which is trained on a large amount of raw text.



DSSM requires labeled pairs for training.
Since there is no labeled data for our tasks, we propose two unsupervised DSSMs, or UDSSMs.
As shown in Figure~\ref{fig:dssm}(b) and \ref{fig:dssm}(c), $(\mathbf{S}^{x}, \mathbf{S}^{y})$ are encoded into contextual representations by deep neural networks $f_1$ and $f_2$; then we compute pair-wise their co-reference scores. 

In what follows, we will describe two assumptions we propose to harvest training data from raw text. 
\textbf{Assumption I: A pronoun refers to one of its preceding nouns in the same sentence.} The sentences generated by this assumption will be used for training  UDSSM-I. Some examples will be shown in the ``data generation" section.
\textbf{Assumption II: In a sentence, pronouns of the same gender and plurality are more likely to refer to the same antecedent than other pronouns.} Similarly, the sentences following the assumption will be used for training UDSSM-II.

Note that the two models, UDSSM-I and UDSSM-II are trained on different types of pair-wise training data, thus the model structures are different, as illustrated in Figure~\ref{fig:dssm}(b) and \ref{fig:dssm}(c), respectively. 
Experiments demonstrated that our methods outperform stat-of-the-art performance on the tasks of WSC and PDP.

%% file: related.tex
\section{Related Work}
\label{sect:related}

As a key component of natural language understanding, commonsense reasoning has been included in an increasing number of tasks for evaluation: COPA~\cite{roemmele2011choice} assesses commonsense causal reasoning by selecting an alternative, which has a more plausible causal relation with the given premise. 
Story Cloze Test (ROCStories, \citeauthor{mostafazadeh-EtAl:2016:N16-1}~\citeyear{mostafazadeh-EtAl:2016:N16-1}) evaluates story understanding, story generation, and script learning by choosing the most sensible ending to a short story. 
JOCI~\cite{zhang2016ordinal} generalizes the natural language inference (NLI) framework~\cite{cooper1996using,Dagan2006RTE,snli:emnlp2015,N18-1101} and evaluates commonsense inference by predicting the ordinal likelihood of a hypothesis given a context. Event2Mind~\cite{rashkin2018event2mind} models stereotypical intents and reactions of people, described in short free-form text. 
SWAG~\cite{zellers2018swag} frames commonsense inference as multiple-choice questions for follow-up events given some context. 
\ReCoRD~\cite{zhang2018record} evaluates a machine's ability of commonsense reasoning in reading comprehension.

Among all these commonsense reasoning tasks, the Winograd Schema Challenge (WSC) and Pronoun Disambiguation Problems (PDP)~\cite{levesque2011winograd} are known as the most challenging tasks for commonsense reasoning.
Although both tasks are based on pronoun disambiguation, a subtask of coreference resolution~\cite{soon2001machine,ng2002improving,peng2016event}, PDP and WSC differ from normal pronoun disambiguation due to their unique properties, which are based on commonsense, selecting the most likely antecedent from both candidates in the directly preceding context. 

Previous efforts on solving the Winograd Schema Challenge and Pronoun Disambiguation Problems mostly rely on human-labeled data, sophisticated rules, hand-crafted features, or external knowledge bases~\cite{peng2015coref,SSS1510295,KR147958}. \citet{D12-1071} hired workers to annotate supervised training data and designed 70K hand-crafted features. 
\citet{Sharma:2015:TAW:2832415.2832433,KR147958,SSS1510295,liu2016combing}
utilized expensive knowledge bases in their reasoning processes. 
Recently, \citeauthor{trinh2018simple}~\citeyear{trinh2018simple} applied neural language models trained with a massive amount of unlabeled data to the Winograd Schema Challenge and improved the performance by a large margin. 
In contrast, our unsupervised method based on DSSM significantly outperforms the previous state-of-the-art method, with the advantage of capturing more contextual information in the data.

%% file: model.tex
\begin{figure*}[t]
\centering
\includegraphics[width=0.99\textwidth]{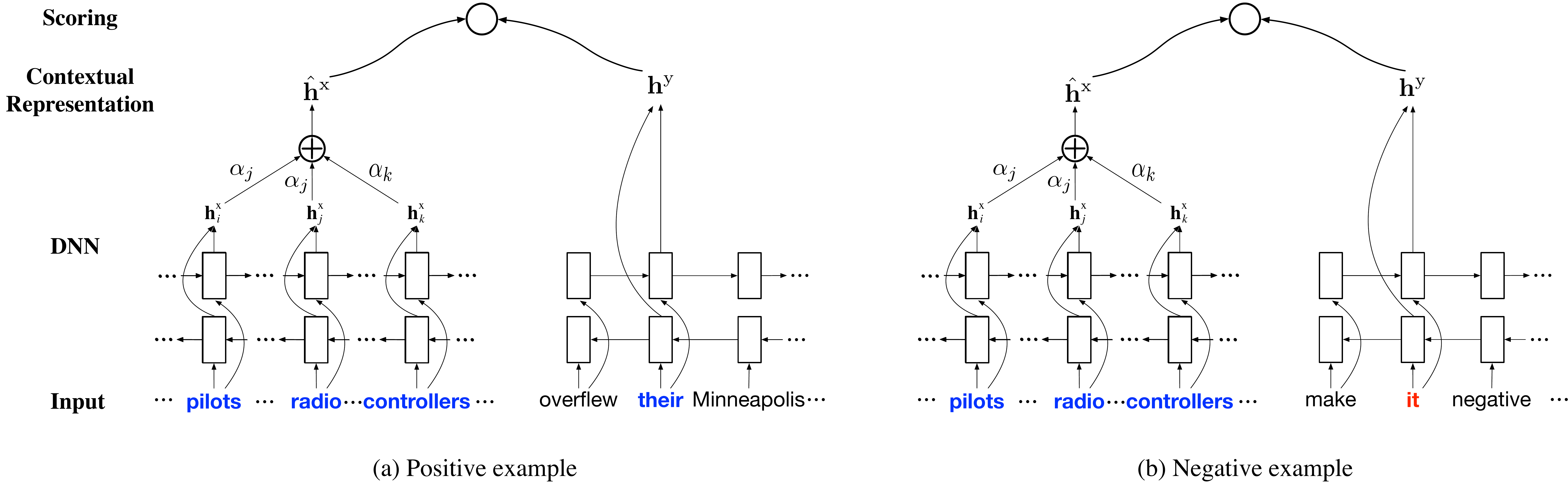}
\label{fig:model}
\caption{The procedure of using UDSSM-I to compute the co-reference scores of a positive example and a negative example respectively. 
The positive example is generated from the sentence `\textit{Two \textbf{Northwest Airlines pilots} failed to make \textbf{radio contact} with \textbf{ground controllers} for more than an hour and overflew \textbf{their} Minneapolis destination by 150 miles before discovering the mistake and turning around.}''. 
The negative one replaces the second sequence with one sequence from different sentence.
}
\label{fig:wdssm-i}
\end{figure*}

\begin{figure*}[t]
\centering
\includegraphics[width=1\textwidth]{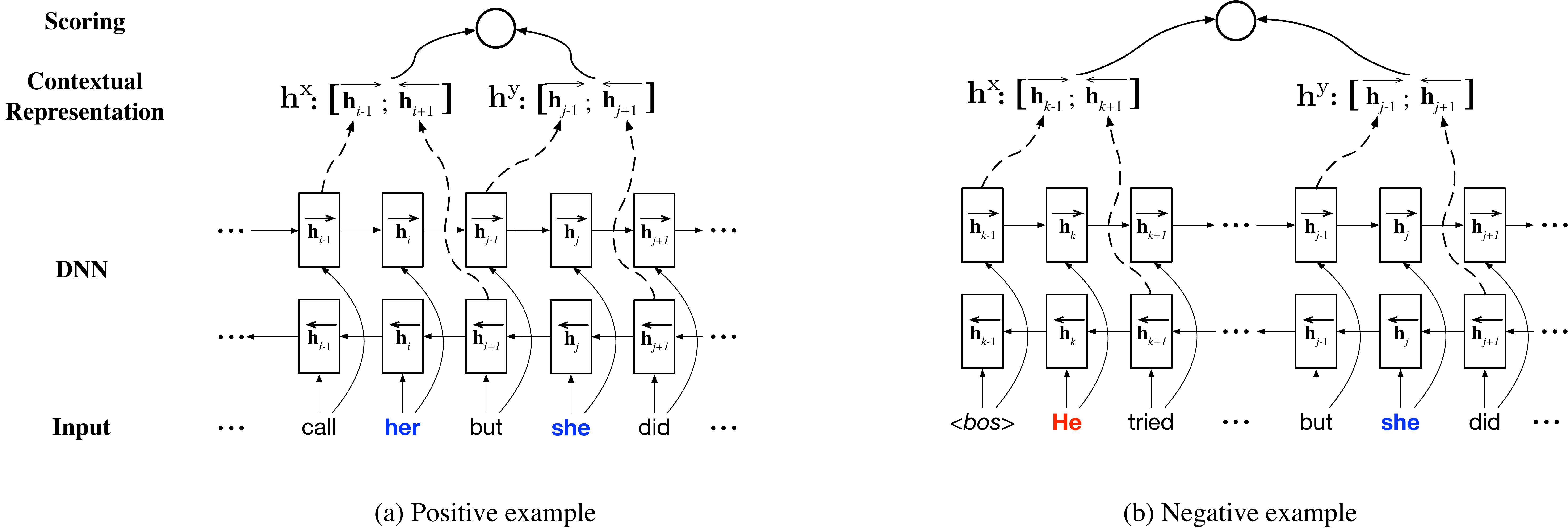}
\label{fig:udssm-sim}
\caption{The procedure of using UDSSM-II to compute the co-reference scores of a positive example and a negative example respectively. Both examples are generated from the sentence ``\textit{He tried twice to call her but she did not answer the phone}''. }
\label{fig:wdssm-ii}
\end{figure*}

\section{Approach} 
\label{sect:modelov}
As shown in Figure~\ref{fig:dssm}, we propose two unsupervised deep structured semantic models~(\textbf{UDSSM-I} and \textbf{UDSSM-II}), which consist of two components: DNN encoding and co-reference scoring. 
For the model UDSSM-I, the co-referred word pairs are automatically learned through an attention mechanism, where the attention weights are the co-reference scores for word pairs. 
For the second model UDSSM-II, we will directly optimize the co-reference score during training. After all, we will get the co-reference scoring function, $\text{Score}_{\theta}(x_i,y)$, to compare the candidate answers in the tasks of PDP/WSC. 
Next, we will show the details of our models trained in an unsupervised way.

In the following sections, we will use uppercase symbols in bold, e.g., $\mathbf{S}^{\text{x}}$, to represent matrices. 
Lowercase symbols in bold, e.g., $\mathbf{h}^{\text{x}}$, represent vectors. 
A regular uppercase symbol, e.g., $S^{x}$, represents a lexical sequence.  
A regular lowercase symbol, e.g., $x_i$ or $y$, represents a token.

\subsection{UDSSM-I Model} 
This model is developed based on Assumption I. 
Its architecture is shown in Figure~\ref{fig:wdssm-i}. 
The sentences generated based on this assumption contain a pronoun $y$ and a set of its preceding nouns $\{x_i,x_j...\}$, which includes the referred word by pronoun.
For example, the sentence in Figure~\ref{fig:wdssm-i}.
As there is no clear label for the co-referred word pairs under this assumption, our model will rank the set of nouns $\{x_i,x_j...\}$ which contains the noun that the pronoun $y$ refers to higher than the set which does not. 
And the co-reference score between words will not be optimized directly during training, but is learned indirectly through the attention mechanism.
We will describe in turn how the training data is generated from raw text, the model architecture, and the co-reference scoring function for the final prediction on the tasks of PDP/WSC. 

\subsubsection{Data Generation}
The main challenge of PDP/WSC tasks is that it has no labeled training data. 
Here we introduce a simple method to collect unsupervised training data by leveraging some linguistic patterns. 
Following Assumption 1, the first hypothesis we make is that ``the pronoun refers to one of the preceding nouns'', which is a common phenomenon in well-written stories or news. 
In this way, we generate ($S^{\text{x}}$, $S^{\text{y}}$) pairs from raw text as follows:

\begin{itemize}[leftmargin=*]
\itemsep-0.2em
\item Parse the sentences in the raw text to obtain entity names, nouns and pronouns.
\item Pick sentences that contain at least one pronoun and multiple nouns preceding it.
\item Split each sentence into two sub-sentences to form a positive pair ($S^{\text{x}}$, $S^{\text{y}}$), where $S^{\text{x}}$ is the first sub-sentence with identified nouns and entity names while $S^{\text{y}}$ is the second sub-sentence with a pronoun.
\item One or more negative pairs are generated from ($S^{\text{x}}$, $S^{\text{y}}$) by replacing $S^{\text{y}}$ with $S^{\text{y}_{\textit{neg}}}$ randomly sampled from other positive pairs.
\end{itemize}

We split the sentence with pronouns and nouns into two sub-sequences separated by the previous word of the pronoun.
Therefore, the example sentence in the Figure~\ref{fig:wdssm-i} can be split into two sub-sentences as shown below: 

\begin{itemize}[leftmargin=*]
\itemsep-0.2em
\item $S^{\text{x}}$: `` \textbf{Two Northwest Airlines pilots} failed to make \textbf{radio contact} with \textbf{ground controllers} for more than an hour and'' 
\item $S^{\text{y}}$: ``overflew \textbf{their} Minneapolis destination by 150 miles before discovering the mistake and turning around''.
\end{itemize}

As the sentences are collected from raw text, the co-reference words are not given.
Our proposed UDSSM-I model will learn the co-reference scoring function through attention mechanism based on the generated sequence pairs. 
Next, we will introduce the details of this model.


\subsubsection{Model Architecture}
This method takes the pair of sequences, ($S^{\text{x}}$, $S^{\text{y}}$), as inputs, and computes similarity between the sequences collected from the same sentence. 
As we hypothesize that one of the nouns in the first sequence and the pronoun in the second are co-referred, we only use the contextual representations of nouns and pronoun to represent the sequences. 
To obtain the contextual representation, we first use a bi-directional LSTM to process these sequences \footnote{We use two different LSTMs to process the sequences $S^{\text{x}}$ and $S^{\text{Y}}$ here. This is to make the negative sampling in Eqn.~(\ref{eqn:loss}) more efficient, so that we can directly use the other representations in the same batch as negative ones.}:
\begin{equation}
\mathbf{H}^{\text{x}} = \text{Bi-LSTM}(\mathbf{S}^{\text{x}}),
\mathbf{H}^{\text{y}} = \text{Bi-LSTM}(\mathbf{S}^{\text{y}}),
\label{eqn:pre-2}
\end{equation}
where $\mathbf{S}^{\text{x}} \in \mathbb{R}^{d\times X}$, $\mathbf{S}^{\text{y}} \in \mathbb{R}^{d\times Y}$ are the word embeddings of the two sequences. 
$d$ is the dimension of the word embeddings. 
$X, Y$ are the lengths of the two sequences. 
$\mathbf{H}^{\text{x}}\in \mathbb{R}^{l\times X}$ and $\mathbf{H}^{\text{y}}\in \mathbb{R}^{l\times Y}$ are the hidden states of bi-directional LSTM.
Our model is task-specifically constructed, so we directly use the hidden state of the first pronoun in the second sequence as its representation:
\begin{equation}
f_2(S^y) \; = \; \mathbf{h}^{\text{y}} \; = \; \mathbf{h}^{\text{y}}_2,
\label{eqn:alpha-1}
\end{equation}
where $\mathbf{h}^{\text{y}}_2\in \mathbb{R}^{\text{l}}$ is the second\footnote{We assign the word just before the pronoun to the second sequence, so the pronoun always appears in the second position of the sequence.} vector from $\mathbf{H}^{\text{y}}$ and it represents the contextual information of the pronoun. Next, we will get the representation of the first sequence. As there are multiple nouns in the first sequence and the pronoun usually refers to only one of them, we use the weighted sum of all the LSTM hidden states of the nouns to represent the sequence, $\hat{\mathbf{h}}^{\text{x}}\in \mathbf{R}^{l}$, as follows:
\begin{eqnarray}
\nonumber
\mathbf{H}^{\text{n}} & = & [ \mathbf{h}^{\text{x}}_{i};  \mathbf{h}^{\text{x}}_{j} ; ...]  \\
\nonumber
\mathbf{\alpha} & = & \text{SoftMax}\left( (\mathbf{W}^{\text{g}} \mathbf{H}^{\text{n}}+\mathbf{b}^\text{g}\otimes \mathbf{e}_N)^\text{T} \mathbf{h}^{\text{y}} \right), \\
f_1(S^x) & = & \hat{\mathbf{h}}^{\text{x}} \; = \; \mathbf{H}^{\text{n}}\mathbf{\alpha},
\label{eqn:alpha-2}
\end{eqnarray}
where $i, j ...$ are the positions of the nouns in the sequence $S^x$ and $[;]$ is the concatenation of two vectors. 
$\mathbf{H}^{\text{n}} \in \mathbb{R}^{l\times N}$ are all the hidden states of the nouns\footnote{We use the toolkit of spaCy in Python for POS and NER, and we will remove the sequences that contain less than 2 nouns. } in $\mathbf{H}^{\text{x}}$ in the sequence.  $N$ is the number of nouns in the sequence. $\mathbf{\alpha}\in \mathbb{R}^N$ is the weights assigned for the different nouns and $\hat{\mathbf{h}}^{\text{x}} \in \mathbb{R}^l$ is the weighted sum of all the hidden states of the nouns. 
$\mathbf{W}^{\text{g}}\in \mathbb{R}^{l\times l}$ and $\mathbf{b}^{\text{g}}\in \mathbb{R}^{l}$ are the parameters to learn; $\mathbf{e}_N\in \mathbb{R}^{\text{N}}$ is a vector of all 1s and it is used to repeat the bias vector $N$ times into the matrix. 
Then we will maximize the similarity of the contextual representations of ($\hat{\mathbf{h}}^{\text{x}}, \mathbf{h}^{\text{y}}$). 
Meanwhile, we also need some negative samples $\mathbf{h}^{\text{y}_{\textit{neg}}}_k $ for $\hat{\mathbf{h}}^{\text{x}}$. 
Then our loss function for this method is constructed:
\begin{equation}
    \textit{L} = - \log \left ( \frac{ \exp \left (  \hat{\mathbf{h}}^{\text{x}}  \mathbf{h}^{\text{y}}  \right ) }{\exp \left ( \hat{\mathbf{h}}^{\text{x}}  \mathbf{h}^{\text{y}} \right ) + \sum_{k}^{K} \exp \left ( \hat{\mathbf{h}}^{\text{x}}  \mathbf{h}^{\text{y}_{\textit{neg}}}_k  \right ) }  \right ),
\label{eqn:loss}
\end{equation}
where $\mathbf{h}^{\text{y}_{neg}}_k \in \mathbb{R}^l$ is the randomly sampled hidden state of pronoun from the sequences not in the same sentence with $S^y$. 

\subsubsection{Co-reference Scoring Function}
Overall, the model tries to make the co-reference states similar to each other. The co-reference scoring function is defined:
\begin{equation}
\text{Score}_{\theta}(x_i,y) = g(\mathbf{h}^{\text{x}}_i, \mathbf{h}^{\text{y}}) = (\mathbf{W}^{\text{g}} \mathbf{h}^{\text{x}}_{i} +  \mathbf{b}^{\text{g}})^{\text{T}}  \mathbf{h}^{\text{y}},
\label{eqn:f_sim}
\end{equation}
where the candidate located at the $i$-th position is represented by its LSTM hidden state $\mathbf{h}^{\text{x}}_{i}$  and the pronoun in the snippet is represented by $\mathbf{h}^{\text{y}}$. 
And the output value of this function for each candidate will be used for the final prediction. 
Next, we will introduce the other unsupervised method.


\subsection{UDSSM-II Model} 
This model is developed based on Assumption II. Its architecture is shown in Figure~\ref{fig:wdssm-ii}. 
As the model is similar to the previous one, we will introduce the details in a similar way.
\subsubsection{Data Generation} 
The second assumption is that ``the pronoun pairs in a single sentence are co-reference words if they are of the same gender and plurality; otherwise they are not." 
Based on this assumption, we can directly construct the co-reference training pairs as follows:
\begin{itemize}[leftmargin=*]
\itemsep-0.2em
\item Parse the raw sentences to identify pronouns.
\item Pick sentences that contain at least two pronouns.
\item The sub-sequence pair with pronouns of the same gender and plurality is labeled as a positive pair; otherwise it is labeled as negative.
\item Replace the corresponding pronoun pairs with a special token ``\textbf{@Ponoun}".
\end{itemize}

Take the following sentence as an example: ``\textbf{He} tried twice to call \textbf{her} but \textbf{she} did not answer the phone." There are three pronouns detected in the sentence,
and we assume that the words $\textbf{her}$ and $\textbf{she}$ are co-reference words, while pairs $(\textbf{she}, \textbf{He})$ and $(\textbf{her}, \textbf{He})$ are not. Thus we can obtain three training examples from the given sentence.
However, in the PDP and WSC tasks, models are asked to compute the co-reference scores between pronoun and candidate nouns, instead of two pronouns. Therefore, we replace the first pronoun in the sentence with a place holder; i.e., a negative training pair is generated by splitting the raw sentence into the following two sub-sequences:

\begin{itemize}[leftmargin=*]
\itemsep-0.2em
\item $S^{\text{x}}$: `` \textbf{@Ponoun} tried twice to call her"  
\item $S^{\text{y}}$: ``but \textbf{she} did not answer the phone."
\item label: Negative
\end{itemize}
and the positive training pair can be generated by the same way:
\begin{itemize}[leftmargin=*]
\itemsep-0.2em
\item $S^{\text{x}}$: `` He tried twice to call \textbf{@Ponoun}"  
\item $S^{\text{y}}$: ``but \textbf{she} did not answer the phone."
\item label: Positive
\end{itemize}
Thus, we could directly train the encoder and co-reference scoring components through the generated training pairs. 


\subsubsection{Model Architecture} 
The previous method, UDSSM-I, follows the task setting of PDP/WSC, and builds the model based on the similarity of the representations between nouns and the pronoun. 
As there is no signal indicating the exact alignment between co-reference words,  the model tries to learn it based on the co-occurrence information from large scale unlabelled corpus.
For the method of UDSSM-II, each representation pair ($\mathbf{h}^{\text{x}}, \mathbf{h}^{\text{y}}$) has a clear signal, $r$, indicating whether they are co-referred or not.
For simplicity, we do not have to split the sentence into two parts.
We first use LSTM to process the sentence as follows:
\begin{eqnarray}
 \overrightarrow{ \mathbf{H} } = \overrightarrow{\text{LSTM}} ([\mathbf{S}^{\text{x}}; \mathbf{S}^{\text{y}}]), 
 \overleftarrow{ \mathbf{H} } = \overleftarrow{\text{LSTM}} ([\mathbf{S}^{\text{x}}; \mathbf{S}^{\text{y}} ]), 
\label{eqn:pre}
\end{eqnarray}
where we can concatenate the word embeddings, $[\mathbf{S}^{\text{x}}; \mathbf{S}^{\text{y}}]$, of two sequences collected under Assumption II.
$\overrightarrow{\text{LSTM}}$ and $\overleftarrow{\text{LSTM}}$ are built in different directions, and $ \overrightarrow{ \mathbf{H} }$, $  \overleftarrow{ \mathbf{H} }$ are the hidden states of the corresponding LSTM. 
Suppose that the pronoun  pair in the sentence are located at the $i$-th and $j$-th positions as shown in the bottom part of Figure~\ref{fig:wdssm-ii}(a). 
We use the hidden states around the pronouns as their contextual representations as follows:
\begin{equation}
 f_1(S^x) = \mathbf{h}^{\text{x}}=\begin{bmatrix}
 \overrightarrow{ \mathbf{h}_{i-1} }\\ 
 \overleftarrow{ \mathbf{h}_{i+1} }
\end{bmatrix},
 f_2(S^y) = \mathbf{h}^{\text{y}}=\begin{bmatrix}
\overrightarrow{ \mathbf{h}_{j-1} }\\ 
 \overleftarrow{ \mathbf{h}_{j+1} }
\end{bmatrix},
\end{equation}
where $\begin{bmatrix} \cdot \\ \cdot \end{bmatrix}$ is the concatenation of all the vectors inside it. Then we further concatenate these representation pair:
\begin{equation}
\mathbf{h}^c=\begin{bmatrix}
 \mathbf{h}^{\text{x}}\\ 
\mathbf{h}^{\text{y}}
\end{bmatrix},
\end{equation}
where  $\mathbf{h}^c\in \mathbb{R}^{4l}$, and it will be the input of loss function with cross entropy as follows:
\begin{eqnarray}
 \nonumber
\textit{L} &=& -r\log\left ( \frac{\exp(  \mathbf{w}^{p} \mathbf{h}^c)}{\exp(  \mathbf{w}^{p} \mathbf{h}^c)  + \exp(  \mathbf{w}^{n} \mathbf{h}^c)} \right ) \\ 
\nonumber
 &-& (1-r)\log\left ( \frac{\exp(  \mathbf{w}^{n} \mathbf{h}^c)}{\exp(  \mathbf{w}^{p} \mathbf{h}^c) + \exp(  \mathbf{w}^{n} \mathbf{h}^c)} \right ),
\end{eqnarray}
where $r\in \{0,1\}$ indicates whether the pronouns at the $m$-th and $n$-th positions should be considered co-reference or not. 
$ \mathbf{w}^{p}\in \mathbb{R}^{4l}$ and $ \mathbf{w}^{n}\in \mathbb{R}^{4l}$ are the parameters to learn.

\subsubsection{Co-reference Scoring Function} 

Similar to the Eqn.(\ref{eqn:f_sim}), for each candidate, we use co-reference scoring function $\text{Score}_{\theta}(x_i,y)$ for the answer selection:
\begin{equation}
\text{Score}_{\theta}(x_i,y) = g(\mathbf{h}^{\text{x}}_i, \mathbf{h}^{\text{y}}) = \mathbf{w}^{p} \begin{bmatrix}
 \overrightarrow{ \mathbf{h}_{i-1} }\\ 
 \overleftarrow{ \mathbf{h}_{i+1} } \\
\overrightarrow{ \mathbf{h}_{j-1} } \\ 
 \overleftarrow{ \mathbf{h}_{j+1} }
\end{bmatrix},
\end{equation}
where $i$ is the position of the candidate in the sentence and $j$ is the position of the pronoun.



%% file: exp.tex
\begin{table*}[!ht]
\centering
\begin{tabular}{lcc}
\toprule
                                               & PDP             & WSC             \\
Co-reference Resolution Tool   &   41.7\% & 50.5                                        \\ 
Patric Dhondt (WS Challenge 2016) &  45.0\% & -\\
Nicos Issak (WS Challenge 2016) & 48.3\% & -\\
Quan Liu (WS Challenge 2016 - winner) & 58.3\% & - \\
Unsupervised Semantic Similarity Method (USSM)             & 48.3\%          & -               \\
Neural Knowledge Activated Method (NKAM)             & 51.7\%          & -               \\
USSM + Cause-Effect Knowledge Base                         & 55.0\%          & 52.0\%          \\
USSM + Cause-Effect + WordNet + ConceptNet Knowledge Bases & 56.7\%          & 52.8\%          \\
USSM + NKAM & 53.3\% & \\
USSM + NKAM + 3 Knowledge Bases &  66.7\% & 52.8\% \\
\midrule
ELMo & 56.7\% & 51.5\% \\
Google Language Model~\cite{trinh2018simple}                             & 60.0\%          & 56.4\%          \\

\textbf{UDSSM-I}           & 75.0\%          & 54.5\%          \\
\textbf{UDSSM-II}       & \textbf{75.0\%} & \textbf{59.2\%} \\
\midrule 
\midrule
Google Language Model (ensemble) & 70.0\%          & 61.5\%          \\
UDSSM-I (ensemble) & 76.7\% & 57.1\% \\ 
UDSSM-II (ensemble) & \textbf{78.3\%} &\textbf{ 62.4\%} \\
\bottomrule
\end{tabular}
\caption{The experiment results on PDP and WSC datasets. We compare our models to Goolge LM trained on the same corpus \footnotemark. }
\label{tbl:exp}

\end{table*}

\section{Experiments}
\label{sect:exp}
\footnotetext{The best models reported in the works of \citet{radford2019language} and  \citet{trinh2018simple} are trained on a much larger corpus from Common Crawl. }

In this section, we will introduce the datasets to train and evaluate our models for commonsense reasoning, the hyper-parameters of our model, and the analysis of our results.

\subsection{Datasets} 
\paragraph{Training Corpus}
We make use of the raw text from Gutenberg~\footnote{\url{http://www.gutenberg.org}}, a corpus offerring over 57,000 free eBooks, and 1 Billion Word~\footnote{\url{https://github.com/ciprian-chelba/1-billion-word-language-modeling-benchmark}}, a corpus of news, to train our model.
We first ignore the sentences that contain less than 10 tokens or longer than 50 tokens.
Then, for the model UDSSM-I, we collect all the sentences with the pronoun before which there're at least two nouns.
For UDSSM-II, we collect all the sentences with at least 2 pronouns.
In total, we collect around 4 million training pairs from each corpus for our proposed method respectively, and we split 5\% as validation set.

\paragraph{Evaluation Dataset} 
We evaluate our model on the commonsense reasoning datasets, Pronoun Disambiguation Problems (PDP)~\footnote{\url{https://cs.nyu.edu/faculty/davise/papers/WinogradSchemas/PDPChallenge2016.xml}} and Winograd Schema challenges (WSC)~\footnote{\url{https://cs.nyu.edu/faculty/davise/papers/WinogradSchemas/WSCollection.xml}}, which include 60 and 285 questions respectively. 
Both of the tasks are constructed for testing commonsense reasoning 
and all the questions from these challenges are obvious for human beings to solve with commonsense knowledge, but hard for machines to solve with statistical techniques.

\subsection{Experimental Setting} 
We use the same setting for both our models. 
The hidden state dimension of a single-directional LSTM is set to be 300. 
We use 300 dimensional GloVe embeddings~\footnote{\url{https://github.com/stanfordnlp/GloVe}} for initialization. 
We use Adamax to optimise the model, set learning rate to be 0.002, dropout rate on all layers are tuned from [0, 0.1, 0.2] and the batch size from [30, 50, 100, 200]. 
For the model UDSSM-I, in one batch, we treat all sequence pairs not from the same sentence as negative cases. 
And it takes around 30 hours on a single K40 GPU to train our models, which are much faster than training a large LM~\cite{jozefowicz2016exploring} taking weeks on multiple GPUs.

\subsection{Experimental Results} 
The experiment results are shown in Table~\ref{tbl:exp}. 
Most of the performance in the top of the Table~\ref{tbl:exp} are the models trained with external knowledge bases, such as Cause-Effect~\cite{liu2016probabilistic}, WordNet~\cite{miller1995wordnet}, ConceptNet~\cite{liu2004conceptnet} knowledge bases. 
Unsupervised Semantic Similarity Method (USSM)~\cite{liu2016combing}  is based on the skip-gram model~\cite{mikolov2013distributed} to train word embeddings， and the embeddings of all the words connected by knowledge bases are optimized to be closer. 
Neural Knowledge Activated Method (NKAM)~\cite{liu2016combing} trained a binary classification model based on whether the word pairs appear in the knowledge base. 
One limitation of these methods is that they rely heavily on the external knowledge bases.
Another limitation is that they just linearly aggregate the embeddings of the words in the context, and that's hard to integrate the word order information.
Instead, our model with LSTM can better represent the contextual information.
Besides, our model don't need any external knowledge bases, and achieve a significant improvement on both of the datasets.

We further compare our models with the unsupervised baselines, ELMo~\cite{peters2018deep} which selects the candidate based on the cosine similarity of the hidden states of noun and pronoun. Another unsupervised baseline, Google Language Model for commonsense reasoning~\cite{trinh2018simple}, which compares the perplexities of the new sentences by replacing the pronoun with candidates.
To make a fair comparison to \citet{trinh2018simple}'s work, we also train our single model on the corpus of Gutenberg only. 
We can see that both of our methods get significant improvement on the PDP dataset, and our UDSSM-II can achieve much better performance on the WSC dataset.
We also report our ensemble model (nine models with different hyper-parameters) trained with both corpus of Gutenberg and 1 Billion Word, and it also achieve better performance than Google Language Model trained with the same corpus. 

Finally, we also compare to the pre-trained Coreference Resolution Tool~\cite{clark2016deep,clark2016improving}\footnote{\url{https://github.com/huggingface/neuralcoref}}, and we can see that it doesn't adapt to our commonsense reasoning tasks and can't tell the difference between each pair of sentences from WSC. In this way, our model can get much better performance.
\subsection{Analysis} 

\begin{table}[!ht]

\begin{tabular}{lp{5.5cm}}
\toprule
WSC 1:     & \textbf{Paul} tried to call George on the phone, but \textbf{he} wasn't successful.                      \\
Ours 1: & \textbf{He} tried to call 911 using her cell phone but that \textbf{he} could n't get the phone to work. \\
\midrule
WSC 2:     & Paul tried to call \textbf{George} on the phone, but \textbf{he} was n't available .                     \\
Ours 2: & He tried twice to call \textbf{her} but \textbf{she} did not answer the phone . \\ 
\bottomrule                      
\end{tabular}
\caption{Comparison of the data from WSC and our training data. Our sentences are retrieved from the UDSSM-II training dataset based on the BM25 value for analysis. The pseudo labels in our training data can help identify the co-references in WSC.}
\label{tbl:pattern}
\end{table}
In this subsection, we will conduct further analysis on the reason that our models work, the benefit of our models comparing to a baseline, and the limitation of our proposed models. 

We have a further analysis on the pair-wise sentences, which we collected for training, to check how our model can work.
We find that some reasoning problems can somehow be converted to the paraphrase problem. 
For example, in Table~\ref{tbl:pattern}, we make use of Lucene Index\footnote{\url{http://lucene.apache.org/pylucene/}} with BM25 to retrieve the similar sentences to the WSC  sentences from our training dataset, and make a comparison.
We can see that these pairs are somehow paraphrased each other respectively.
For the first pair, the contextual representations of ``Paul" and ``he" in WSC could be similar to the contextual representations of "he" in our training sentence.
As these representations are used to compute the co-reference score, the final scores would be similar.
The pseudo label ``positive" for our first sentence will make the positive probability of the golden co-references ``Paul" and ``he" in WSC higher.
And for the second pair in Table~\ref{tbl:pattern}, the pseudo label of positive in our second sentence will make the positive probability of the golden co-references ``George" and ``he" in WSC 2 higher.
In this way, these kinds of co-reference patterns from training data can be directly mapped to solve the Winograd Schema Challenges.

Here's another example from PDP demonstrating the benefit of our method: ``Always before, Larry had helped Dad with his work. But \textit{he} could not help him now, for Dad said that …". 
\citet{trinh2018simple} failed on this one, probably because language models are not good at solving long distance dependence, and tends to predict that ``{he} " refers to ``his" in the near context rather the correct answer ``Larry".
And our model can give the correct prediction.

We further analysis the predictions of our model. 
We find that some specific commonsense knowledge are still hard to learn, such as the following pairs:
\begin{itemize}[labelindent=0.5em,labelsep=0.2cm,leftmargin=0.4cm]
\itemsep-0.2em
\item The trophy doesn't fit into the brown \textbf{suitcase} because \textbf{it} is too small.
\item The \textbf{trophy} doesn't fit into the brown suitcase because \textbf{it} is too large.
\end{itemize}
To solve this problem, the model should learn the knowledge to compare the size of the objects.
However, all of our models trained with different hyper-parameters select the same candidate as the co-referred word for ``it" in both sentences.
To solve the problem, broader data need to collect for learning more commonsense knowledge.